\documentclass[twoside,leqno,twocolumn]{article}

\usepackage{SDM_style}

\usepackage{amsfonts,bm}
\usepackage{algorithm}
\usepackage{algorithmic}

\usepackage{booktabs}
\usepackage{amsmath}
\usepackage{graphicx}
\usepackage{subfigure}
\usepackage[hyphens]{url}
\usepackage{amsmath, cite}
\usepackage{multirow}
\usepackage{color}
  \usepackage{dsfont}
\usepackage{braket}
 \usepackage{mathtools}
 \usepackage[hyphens]{url}
 \usepackage{caption}

 \usepackage{enumerate}
 \usepackage{enumitem}
 \usepackage{empheq}
\usepackage{calc}

\begin{document}


\title{\Large Discovery of Shifting Patterns in Sequence Classification}



\author{Xiaowei Jia, Ankush Khandelwal, Anuj Karpatne, Vipin Kumar\\
\footnotesize\baselineskip=9pt Department of Computer Science and Engineering, University of Minnesota\\
\footnotesize\baselineskip=9pt jiaxx221@umn.edu, \{ankush,anuj,kumar\}@cs.umn.edu
}


\date{}
\maketitle

\begin{abstract}
In this paper, we investigate the multi-variate sequence classification problem from a multi-instance learning perspective. Real-world sequential data commonly show discriminative patterns only at specific time periods. For instance, we can identify a cropland during its growing season, but it looks similar to a barren land after harvest or before planting. 
Besides, even within the same class, the discriminative patterns can appear in different periods of sequential data. Due to such property, these discriminative patterns are also referred to as shifting patterns. The shifting patterns in sequential data severely degrade the performance of traditional classification methods without sufficient training data. 

We propose a novel sequence classification method by automatically mining shifting patterns from multi-variate sequence. The method employs a multi-instance learning approach to detect shifting patterns while also modeling temporal relationships within each multi-instance bag by an LSTM model to further improve the classification performance. 
We extensively evaluate our method on two real-world applications - cropland mapping and affective state recognition. The experiments demonstrate the superiority of our proposed method in sequence classification performance and in detecting discriminative shifting patterns.
\end{abstract}

\section{Introduction}
The last decade has witnessed the rapid development of Internet and sensor equipment, which produce large volume of sequential data. The collected sequential data usually contain descriptive information from multiple aspects, which form multi-variate data streams. For instance, the optical satellite sensors can capture reflectance values for multiple bandwidths, which are indicative of different environmental variables, such as vegetation, aerosols, water index, etc. The classification of these sequential data is of great importance in many applications. For example, the cropland mapping using multi-spectral remote sensing data can offer timely agricultural information, which is critical to meet the increasing demand for food supply and food security.  



\begin{figure} [!t]
\centering
\subfigure[]{ \label{fig:a}
\includegraphics[width=0.3\columnwidth]{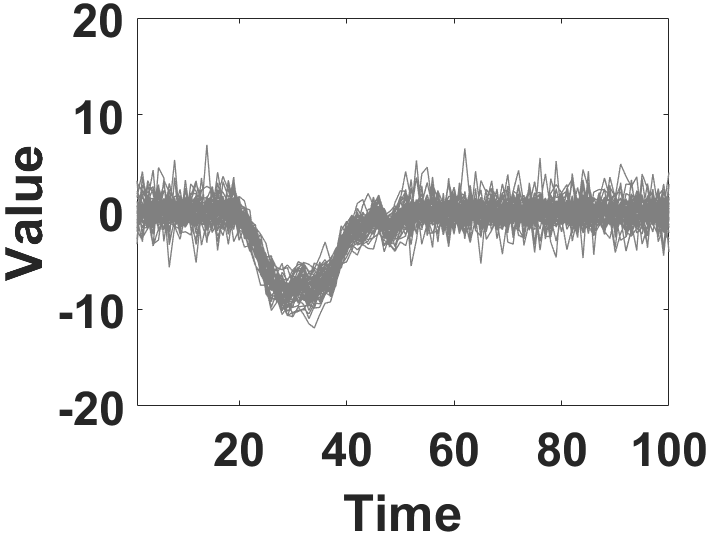}
}\hspace{-.15in}
\subfigure[]{ \label{fig:b}
\includegraphics[width=0.3\columnwidth]{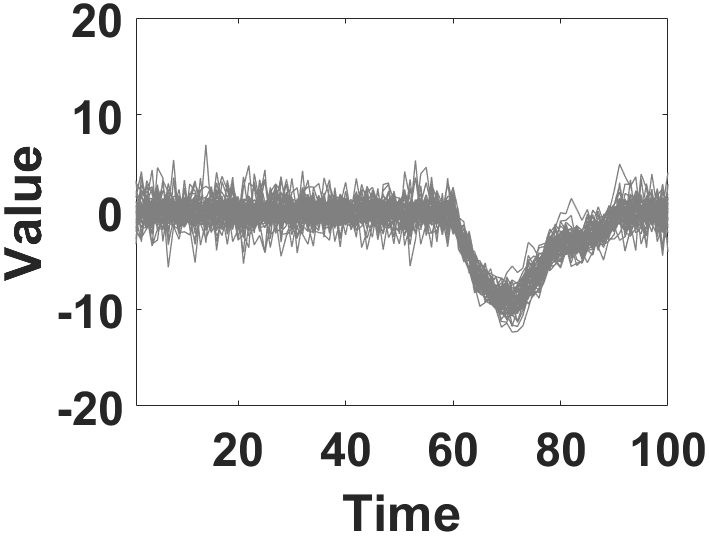}
}\hspace{-.15in}
\subfigure[]{ \label{fig:b}
\includegraphics[width=0.3\columnwidth]{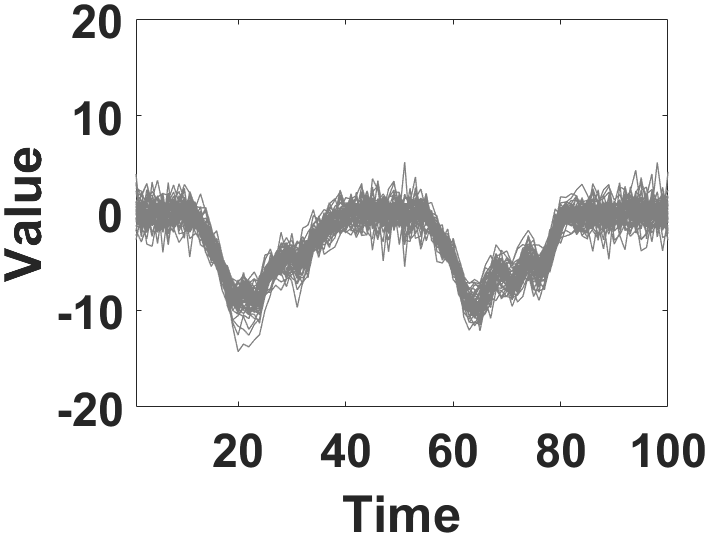}
}\vspace{-.2in}
\caption{Example vegetation sequences of the same class (burned area) with shifting patterns.}
\label{toy}
\end{figure}

Many sequential datasets are collected over long span of time and contain much irrelevant information to the classification task. When classifying multi-variate sequential data, each class is usually reflected by certain discriminative patterns within the sequence. Even for sequences within the same class, the discriminative patterns can appear in different time periods. 
Due to such property, we call these discriminative periods as \textit{shifting patterns}. Consider the three example sequences of the same class in Fig.~\ref{toy}. We can observe similar patterns in different periods and even with different number of occurrences. Therefore, the effective classification requires the automatic detection of the most informative period from the entire sequence~\cite{cheng2007discriminative}. 



Shifting patterns are ubiquitous in real-world applications. In cropland mapping, domain experts commonly refer to growing season to characterize specific crop types. However, the planting and harvest time can vary across regions and across years according to climate conditions. 
Similarly, 
when electroencephalogram (EEG) data are collected for affective state recognition, 
the emotion signature is only reflected in short time periods and these informative periods also change across users. Given multi-variate sequential data collected in these applications, the discriminative patterns are often more complicated than the patterns in uni-variate sequences (e.g. Fig.1) and the data also contain much noise. Hence, the shifting patterns cannot be easily recognized using simple time series-based methods~\cite{verbesselt2010detecting,chen2016general}. 

Most traditional sequence classification approaches directly take the input of entire sequence and treat each time step equally without the awareness of the informative period.
Hence their performance is highly likely to be negatively impacted by the irrelevant time periods in sequential data. Also, the ignorance of the shifting pattern property can result in strong heterogeneity/multi-modality in training set. Consequently, the heterogeneity will lead to 
poor learning performance without sufficient labeled data on all the modes. In contrast, the successful discovery of shifting patterns can not only improve the classification performance, but also promote the interpretability of classification. In addition, the successful detection of informative period in streaming data can potentially contribute to an early-stage prediction before collecting the entire sequence.

There exist several research works that explore pattern mining in sequence classification~\cite{wang2005harmony,xun2016detecting,lee2011mining,yuan2011discriminative,batal2012mining}. For instance, Xun \textit{et al.}~\cite{xun2016detecting} adopt a dictionary learning method to detect frequent patterns and then transform sequential data into a pattern-based representation for classification. 
Wang \textit{et al.}~\cite{wang2005harmony} propose a rule-based classifier HARMONY that discovers the strong associations between frequent patterns and class labels. Rather than detecting discriminative patterns, these works are mostly focused on frequent patterns, which are likely to reflect common non-discriminative patterns shared among classes and also vulnerable to noise. Some other works~\cite{yuan2011discriminative,lee2011mining} utilize predefined or constrained discriminative patterns in classification, but they lack the ability to automatically extract discriminative patterns.

In this paper, we propose a sequence classification method, which integrates a multi-instance learning (MIL) approach to detect discriminative patterns within multi-variate sequences. 
In particular, we introduce a sliding time window to capture different time periods over the sequence, 
which is further combined with a Long-Short Term Memory (LSTM) to model temporal dependencies in sequential data. 
LSTM has shown tremendous success in memorizing temporal dependencies in long-term events, which commonly exist in 
climate changes, health-care, 
etc. By incorporating temporal relationships in sequential data, LSTM assists in better uncovering discriminative patterns within each time window. For example, crops in their mature phase are similar to certain tree plantations, but we can better distinguish between them by combining the growing history at previous time steps. Besides, the modeling of temporal dependencies assists in mitigating the impact of noise and outliers.

Combining the information from both current time period and history, each time window generates a latent output indicating the detection confidence of discriminative patterns. After gathering the latent outputs from all the time windows, we utilize an MIL approach to aggregate them 
into final sequence label. Compared with traditional supervised approach, MIL approach learns a mapping from a bag of instances (i.e. time windows) to a label on the whole bag (i.e. sequence). 
However, in many existing MIL approaches, different instances are treated independently. In contrast, the LSTM-based sequential structure in our method models the temporal relationships among instances in the bag. In this way, each instance interacts with other instances in the sequential order and contributes to their latent outputs.  
In addition, we 
will discuss using context information to improve the performance of MIL approach.

We extensively evaluate the proposed method in two real-world applications - cropland mapping using remote sensing data and affective state recognition using EEG data. Cropland mapping is challenging for agricultural domain researchers because different crop types look similar in most dates and are only distinguishable in certain periods of a year. In the second task, affective states are reflected by short neural activities and the classification becomes even more difficult 
without large volume of labeled sequences. The experimental results confirm that our proposed method outperforms multiple baselines in both tasks. In addition, we demonstrate that the proposed method can successfully detect shifting patterns and provide reasonably good performance in early-stage prediction. 



\section{Problem Definition}
In this work, we are given a set of $N$ data points, 
$Z=\{z_1,z_2,...,z_N\}$. Each sample $z_i$ is a sequence of multi-variate features at $T$ time steps, $z_i=\{z_i^1,...,z_i^T\}$, where 
$z_i^t\in\mathbb{R}^D$. Also, we are provided with the labels of these sequential data, $Y=\{y_1,y_2,...,y_N\}$. In method discussion, we omit the sample index $i$ when we focus on a single data point and cause no ambiguity. 

Our objective is to train a classification model using the provided sequential data and labels. The learned model can then be applied to predict the label for any test sequences. For instance, in croplands mapping, we train the model using the labels for certain regions in a specific year. Then we utilize the learned model to detect target crop types in other regions or in other years. 
In addition, we aim to 
locate the most discriminative time period for each sequence sample. 


\section{Method} \label{s_method}


\subsection{Sequential classification model}
Sequential data in real applications are often collected over a long span of time, and therefore cover many time periods irrelevant to the classification task. 
While the class information can be reflected by discriminative shifting patterns, in practice we are usually not aware of the informative time periods in advance. 

In this paper, we propose to discover the discriminative shifting patterns from sequential data and subsequently leverage the detected patterns for classification. 
Specifically, to detect discriminative patterns that may appear at any position in a given sequence, we introduce a sliding time window with length $w$. The basic intuition is to move this sliding window along the sequence and utilize the sliding window to capture the informative time periods that reflect the discriminative patterns. It is noteworthy that the length of the sliding window $w$ depends on specific applications. For instance, in EEG monitoring, $w$ depends on the time span of neural activities (more details will be provided in Section~\ref{sec:exp}). 
In method discussion, we set the step size (i.e. stride) of sliding window to be 1. In practice, 
we can also increase the step size to reduce computational cost.

\begin{figure} [!h]
\centering
\includegraphics[width=0.85\columnwidth]{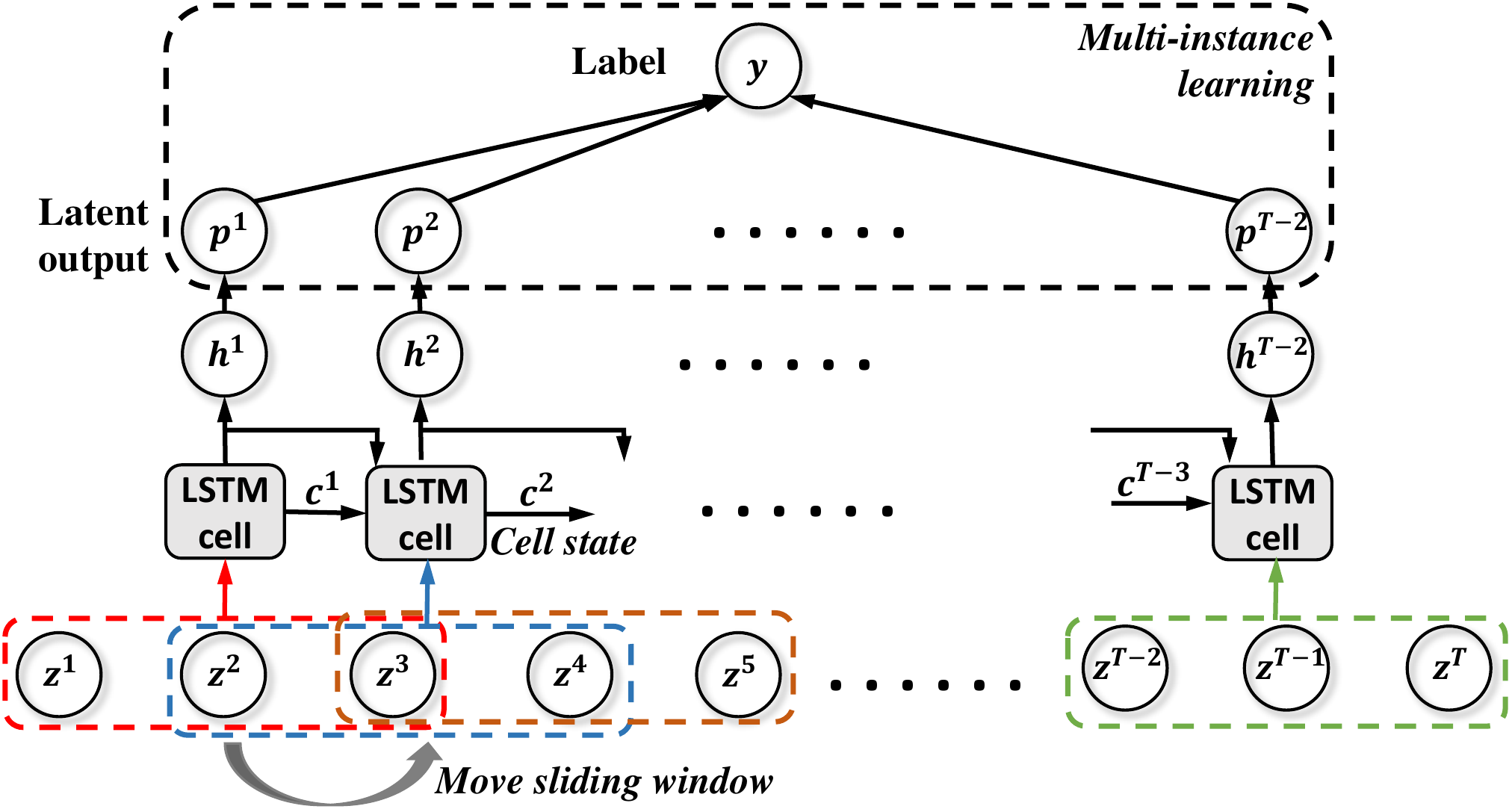}\vspace{-.12in}
\caption{The proposed sequential classification method: the raw features in each sliding time window is fed into LSTM structure, and the outputs from LSTM structure are aggregated by an MIL approach.}
\label{flow_chart}
\end{figure}

While moving the sliding window along the sequence, we generate a latent output $p^t$ for each time window $[t,t+w-1]$. Here $p^t$ represents the detection confidence of discriminative patterns for $K$ different classes in this time window. Besides, we model the temporal dependencies between different time windows using Long-Short Term Memory (LSTM), as shown in Fig.~\ref{flow_chart}. We utilize LSTM mainly because of its capacity to memorize long-term events, which are very common in real-world scenarios, including climate changes, emotion monitoring, disease propagation, etc. 

In this way, we capture the local patterns within each time window by learning the mapping from the time window to a hidden representation (through LSTM cell) while utilizing LSTM to model the global temporal patterns over a long period. Then we will aggregate latent outputs from all the time windows to generate final classification result $y$ for the sequence via an MIL method. The proposed MIL structure enables modeling the contribution of each time window to the final decision making.

In this section, we first introduce the proposed LSTM-based sequential model. After that we provide details on the MIL method to combine multiple latent outputs. Finally we discuss how to integrate context information to further improve the classification.  

\subsubsection{LSTM-based sequential model}
For each time window that starts from time $t$, we represent the raw input features within the time window as $x^{{t}}=\{z^t,z^{t+1},...,z^{t+w-1}\}$. Hereinafter we utilize the staring time step $t$ as the index of a time window. As each time window $t$ contains multi-variate features at several consecutive time steps, we wish to extract more representative local patterns from each time window $x^t$, thus to felicitate distinguishing between different classes. Therefore, we introduce the hidden representation $h^{{t}}$. As observed from Fig.~\ref{flow_chart}, $h^{{t}}$ is generated via an LSTM cell using both the raw features $x^{t}$ within the current time window and the information from previous time window. Then the latent output $p^t$ is generated from $h^t$.

The discriminative patterns of each class usually follow specific temporal evolutionary process. For example, during certain period, the sequential data may gradually show stronger signal of discriminative pattern, and then the signal strength gradually decreases. 
The LSTM structure in Fig.~\ref{flow_chart} is capable of modeling the temporal evolution, which improves the classification performance and also assists in alleviating the impact of noise at any individual time steps.

\begin{figure} [!h]
\centering
\includegraphics[width=0.7\columnwidth]{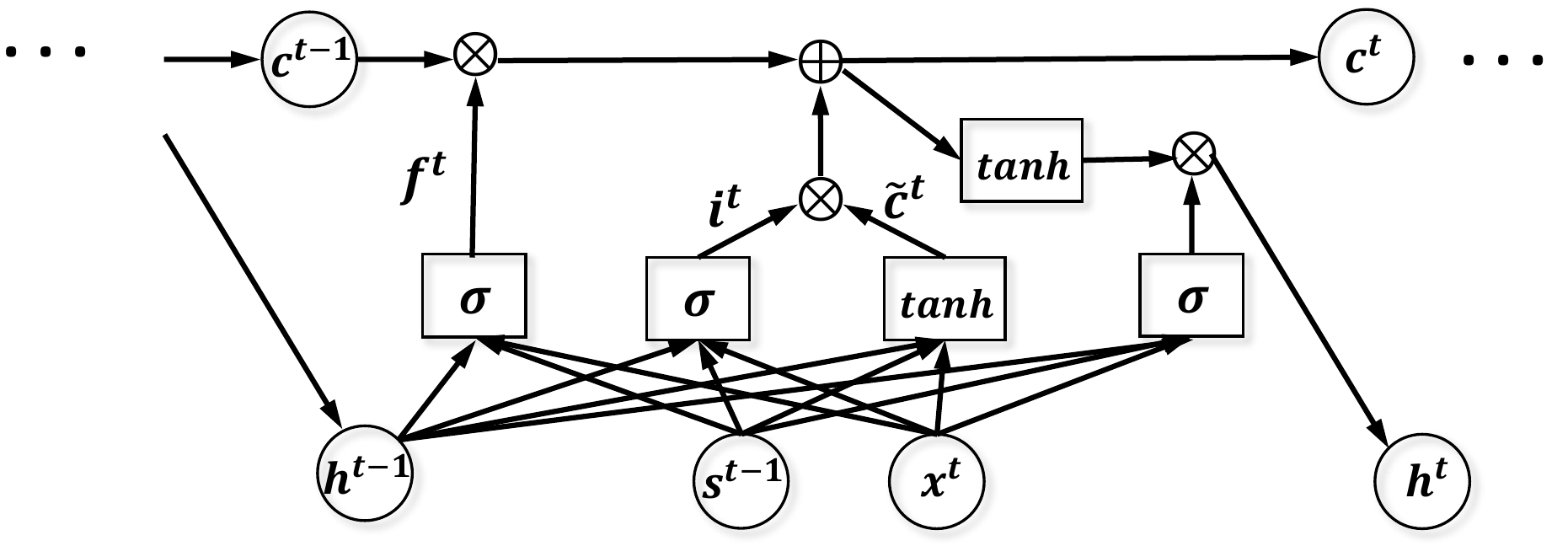}\vspace{-.12in}
\caption{The structure of LSTM cell.}
\label{lstm_cell}
\end{figure}


We now briefly introduce the LSTM cell, as shown in Fig.~\ref{lstm_cell}. Each LSTM cell contains a cell state $c^t$, which serves as a memory and allows the hidden units $h^t$ to reserve information from the past. 
The cell state $c^t$ is generated by combining $c^{t-1}$ and the information at $t$. Hence the transition of cell state over time forms a memory flow, which enables the modeling of long-term dependencies. Specifically, we first generate a new candidate cell state $\tilde{c}^t$ by combining $x^t$ and $h^{t-1}$ into a $tanh(\cdot)$ function, as:
\begin{equation}
\footnotesize
\begin{aligned}
\tilde{c}^t &= tanh(W^c_h h^{t-1} + W^c_x x^t),
\end{aligned}
\end{equation}
where $W^c_h\in \mathds{R}^{H\times H}$ and $W^c_x\in \mathds{R}^{H\times Dw}$ denote the weight parameters used to generate candidate cell state. Hereinafter we omit the bias terms as they can be absorbed into weight matrices. 
Then a forget gate layer $f^t$ and an input gate layer $g^t$ are generated using sigmoid functions, as follows:
\begin{equation}
\footnotesize
\begin{aligned}
f^t &= \sigma(W^f_h h^{t-1} + W^f_x x^t),\\
g^t &= \sigma(W^g_h h^{t-1} + W^g_x x^t),
\end{aligned}
\end{equation}
where \{$W^f_h$, $W^f_x$\} and \{$W^g_h$, $W^g_x$\} denote two sets of weight parameters for generating forget gate layer $f^t$ and input gate layer $g^t$, respectively.
The forget gate layer is used to filter the information inherited from $c^{t-1}$, and the input gate layer is used to filter the candidate cell state at time $t$. In this way we obtain the new cell state $c^t$ as follows:
\begin{equation}
\footnotesize
c^t = f^t\otimes c^{t-1}+g^t\otimes\tilde{c}^t,
\end{equation}
where $\otimes$ denotes entry-wise product.


Finally, we generate the hidden representation at $t$ by filtering the the obtained cell state using a output gate layer $o^t$, as:
\begin{equation}
\footnotesize
\begin{aligned}
o^t &= \sigma(W^o_h h^{t-1} + W^o_x x^t),\\
h^t &= o^t\otimes tanh(c^t),
\end{aligned}
\label{hgen}
\end{equation}
where $W^o_h\in \mathds{R}^{H\times H}$ and $W^o_x\in \mathds{R}^{H\times Dw}$ are the weight parameters used to generate the hidden gate layer.

With the hidden representation $h^t$, we produce the latent output of each time window ${t}$ using a sigmoid function with parameter $U\in \mathds{R}^{K\times H}$, as follows:
\begin{equation}
\footnotesize
\begin{aligned}
p^{{t}}&=\sigma(Uh^{{t}}).
\end{aligned}
\label{pgen}
\end{equation}
The parameters in the LSTM model can be estimated by back-propagation (BP) algorithm. 

\subsubsection{Multi-instance aggregation layer}
Having obtained the latent outputs, we introduce an MIL approach, which aims to establish a mapping from multiple time windows to the label of entire sequence. Specifically, the MIL structure aggregates the latent outputs obtained from sliding time windows, \{$p^1$, $p^2$,..., $p^{T-w+1}$\}. Since this series of latent outputs indicates the temporal evolution of discriminative knowledge as the sliding window moves along the sequence, we call this series as the \textit{temporal profile}.

\begin{figure} [!h]
\centering
\includegraphics[width=0.82\columnwidth]{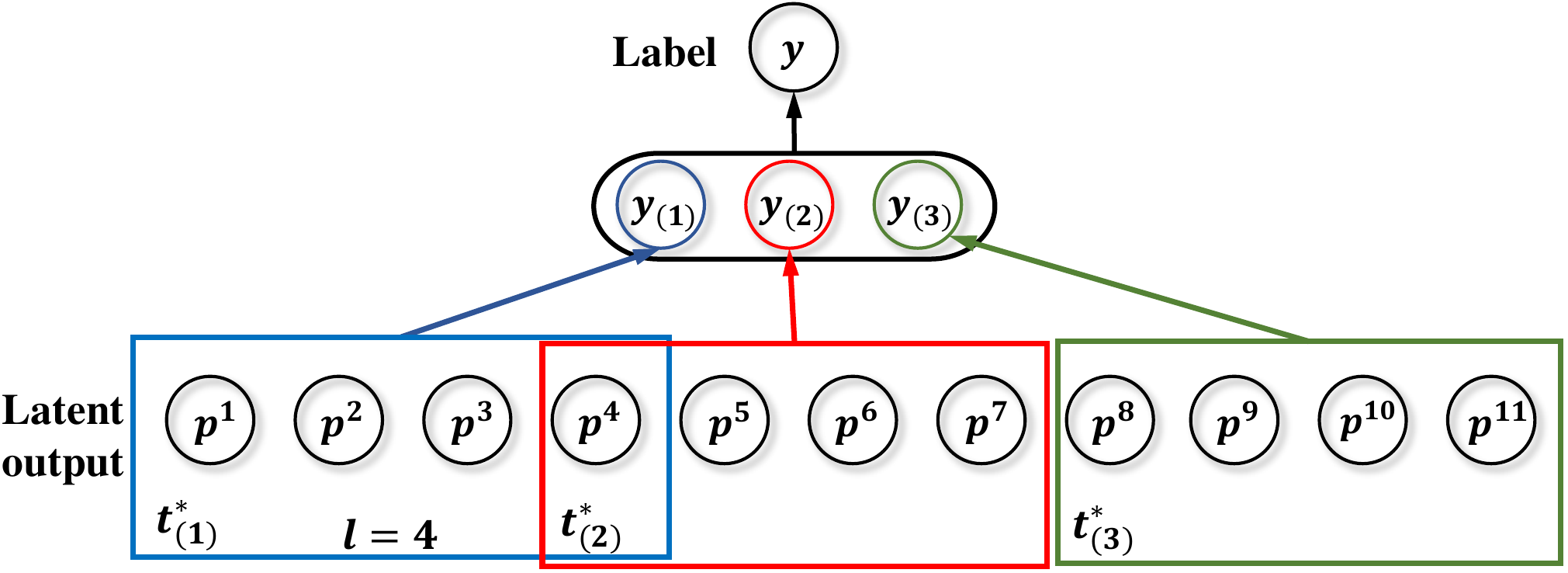}\vspace{-.12in}
\caption{The MIL approach to aggregate $p^t$ from multiple time windows. In this example, $l_k=4$ for $k=1$ to $3$. The discriminative patterns of class $k$ persist for $l_k$ consecutive time windows/latent outputs, shown by the boxes in blue, red and green color, respectively.}
\label{multi_instance}
\end{figure}

According to the property of shifting patterns, if there exists one time window that shows strong discriminative pattern of a specific class, then the sequence should belong to this class. However, the ubiquitous noise in sequential data frequently leads to misclassification, 
which likely disturbs the latent outputs from time windows. To address this issue,  
we assume that the discriminative pattern should persist for several consecutive time windows, as shown in Fig.~\ref{multi_instance}. Such assumption conforms to most real scenarios, and also mitigates the impact of noise and outliers at individual time steps. 

For each class $k\in[1,K]$, instead of selecting the time window with largest $p^t$ value, we take the maximum of the average $p^t$ value over consecutive $l_k$ time windows:
\begin{equation}
\footnotesize
y_{(k)} = \max_{{t}}\,avg(p^{t}_{(k)},p^{t+1}_{(k)},...,p^{t+l_k-1}_{(k)}), k=1,...,K,
\label{i2}
\end{equation}
where $p^t_{(k)}$ denotes the $k^{th}$ entry of $p^t$. We can observe that the larger $y_{(k)}$ requires the higher average value of $p^t_{(k)}$ for consecutive $l_k$ time windows. 

Then we adopt a soft-max function to generate posterior probability for each class $k$:
\begin{equation}
\footnotesize
P(\hat{y}=k|x)=\frac{exp(y_{(k)})}{\sum_{k'}exp(y_{(k')})},
\label{i3}
\end{equation}
where we utilize $\hat{y}$ to distinguish between the predicted label and the provided label $y$. 

Note that we set $l_k$ separately for each class $k$ since different classes can have different length of discriminative periods, e.g. long-season crops vs. short-season crops. In this work, we propose a self-adaptive method to adjust the value of $l_k$ for each class $k$. During each update iteration in BP, we can obtain $p^{t=1:T-w+1}_k$ through feed-forward process. We utilize $P_k$ to represent the distribution of $p_k$ values across all the time windows from $1$ to $T-w+1$ and over all the training samples in class $k$. Then we compute the average $p^{t=1:T-w+1}_k$ over all the samples in class $k$, denoted by $\varphi^{t=1:T-w+1}_k$. For each class $k$, we select $l_k$ to be sufficiently long to cover the consecutive time windows with stronger discriminative signals than the remaining periods. In our tests, we set $l_k$ to be the maximum number of consecutive time windows, s.t. $\exists t'$, for $t=t'$ to $t'+l_k-1$, $\varphi_k^t$ is larger than 80 percentile of $P_k$ .



\subsubsection{Context filtering}
One limitation of the aforementioned MIL approach lies in its vulnerability to temporally auto-correlated noise. For instance, the collected sequential data may contain much noise in a long period because of data acquisition errors. In this case, the noise affects the latent outputs for several consecutive time windows, 
likely resulting in misleading outcomes by Eqs.~\ref{i2} and~\ref{i3}. 

To tackle this problem, we further incorporate the context information, which is commonly available in real-world applications. The context information describes a clustering structure of training samples, which can be determined by geo-spatial information, data source properties, etc. The samples in the same cluster usually share similar temporal profiles. 
Considering croplands in the same region, 
farmers are prone to planting and harvesting each crop type in close dates because of 
climate conditions. 
In EEG monitoring, the collected EEG sequences from users with the same experience, e.g. watching the same videos, are likely to share similar temporal profiles. By incorporating the context information in training process, we wish to collaboratively fix the latent outputs disturbed by temporally auto-correlated noise. In this work, we assume that the context information is available in applications or already provided by domain researchers. 

Assuming there are in total $M$ different contexts (i.e. $M$ clusters of samples), 
we define a mapping $\mathcal{C}(i)$ from each sequence index $i$ to its context index in $[1,M]$. Then based on the proposed sequential model, we embrace the context knowledge as a regularization term in the cost function. Considering a sequence $i$, for each class $k$, we aim to regularize the temporal profile $p_{i,(k)}$ to stay close to the average temporal profile $\bar{p}_{\mathcal{C}(i),(k)}$ over all the samples in class $k$ and in the same context $\mathcal{C}(i)$. 
More formally, the entire cost function can be expressed as follows:
\begin{equation}
\footnotesize
\begin{aligned}
\mathcal{J}\!=\!\sum_i\sum_k\!\{-\!\mathds{1}(y_i=k)log(p(\hat{y_i}=k|x)\!+\!\!\lambda 
\sum_t (p^{t}_{i,(k)}-\bar{p}^{t}_{\mathcal{C}(i),(k)})^2\},
\end{aligned}
\label{cost}
\end{equation}
where 
$\mathds{1}(\cdot)$ denotes the indicator function, and $\lambda$ is the weight of the regularizer. The first term on the right side represents the entropy-based cost for soft-max function, and the second term is adopted to regularize samples in the same context to share similar temporal profiles.

We compute the gradient with respect to $p_{(k)}^{t}$ as follows. Again we omit the sequence index $i$ and context index $\mathcal{C}(i)$ for simplicity.
\begin{equation}
\footnotesize
\begin{aligned}
&\frac{\partial \mathcal{J}}{\partial p^t_{(k)}}\!=\!  
\begin{cases}
p(\hat{y}=k|x)\!-\!\mathds{1}(y=k)\!+\!2\lambda (p^t_{(k)}\!-\!\bar{p}_{(k)}^t),\hspace{-.07in}&\!t\in [t_{k}^*, t_{k}^*\!+\!l_k\!-\!1] \\
2\lambda (p^t_{(k)}\!\!-\!\bar{p}_{(k)}^t),&\!\text{otherwise},
\end{cases}\\
&\text{where}\,\, t_{k}^* = \arg\!\max_t avg(p^{t}_{(k)},p^{t+1}_{(k)},...,p^{t+l_k-1}_{(k)}).
\end{aligned}
\label{gd}
\end{equation}

The gradient of $p^t_{(k)}$ w.r.t. model parameters 
can be estimated by BP algorithm. The complete learning process is summarized in Algorithm 1. 
The time complexity is $O(NTKd)$, where the number of classes $K$ is a constant factor in most cases. $d$ is a constant factor determined by the dimension of input features, hidden representation and number of different contexts. 
We name the proposed method as \textbf{S}hifting \textbf{P}attern \textbf{A}nalysis from \textbf{M}ulti-variate \textbf{S}equences (SPAMS).



\begin{algorithm}[!t]
\footnotesize
\caption{Learning process for SPAMS.}
\begin{algorithmic}[1]
\REQUIRE $\{z_i^1,z_i^2,...,z_i^T\}_{i=1}^N$: a set of multi-variate sequences.\\
$\{y_1,...,y_N\}$: labels of sequences. $\mathcal{C(\cdot)}$: context information.
    \STATE Initialize parameters.
    \WHILE{not converge \&\& $it\text{++} \le MaxIter$}
    \FOR {$i \gets 1$ \textbf{to} $N$}
    \STATE// \textit{In practice we use mini-batch update}.
    \FOR {$t \gets 1$ \textbf{to} $T-w+1$} 
    \STATE Compute latent output $p_i^t$ by Eqs.~\ref{hgen} and~\ref{pgen}.
    \ENDFOR
    \FOR{$k \gets 1$ \textbf{to} $K$} 
    \STATE Find $t^*_{i,(k)}$ based on the computed $p_i^{t=1:T-w+1}$.
    \ENDFOR
    \STATE Compute the predicted label $\hat{y}$ by Eqs.~\ref{i2} and~\ref{i3}.
    \STATE Update parameters by Eq.~\ref{gd} and back-propagation.
    \ENDFOR
    \FOR{$k \gets 1$ \textbf{to} $K$}
    \STATE Update $l_k$ based on latent outputs.
    \STATE Compute $\bar{p}_{(k)}$ for $M$ contexts using obtained $\{p_i\}_{i=1}^N$.
    \ENDFOR
    \ENDWHILE
\end{algorithmic}
\label{alg2}
\end{algorithm}

In practice, given incoming data streams, we are usually interested in prediction at early stage. For instance, the government would like to identify local crop types at the growing season rather than at the end of each year. 
Using the proposed method, we can obtain the latent output/detection confidence of each time window ${t}$, which can be treated as the classification result of the corresponding time period $[t,t+w-1]$. If short delay is allowed, e.g. predicting for time window $[t-w+1,t]$ at time step $t+l-1$, we can also compute the posterior probability (Eqs.~\ref{i2} and~\ref{i3}) up to $t+l-1$, which is more resistant to noise.  

\begin{table*}[!t]
\footnotesize
\newcommand{\tabincell}[2]{\begin{tabular}{@{}#1@{}}#2\end{tabular}}
\centering
\caption{Performance($\pm$standard deviation) of each method in cropland mapping using AUC and F-1 score.}
\begin{tabular}{l|cc|cc|cc}
\hline
 &\multicolumn{2}{c|}{\textbf{\textit{R1}}} & \multicolumn{2}{c|}{\textbf{\textit{R2}}} & \multicolumn{2}{c}{\textbf{\textit{Cross-year}}} \\
\textbf{Method} & \textbf{AUC} & \textbf{F1} & \textbf{AUC} & \textbf{F1} & \textbf{AUC} & \textbf{F1} \\ \hline 
ANN & 0.717($\pm$0.018) & 0.711($\pm$0.011) & 0.715($\pm$0.018) & 0.704($\pm$0.012) & 0.660($\pm$0.021) & 0.678($\pm$0.015)\\
RF & 0.746($\pm$0.013) & 0.733($\pm$0.010) & 0.744($\pm$0.013) & 0.734($\pm$0.012) & 0.655($\pm$0.014) & 0.678($\pm$0.012)\\
SVM$^{hmm}$ & 0.753($\pm$0.015) & 0.737($\pm$0.011) & 0.746($\pm$0.016) & 0.733($\pm$0.011) & 0.721($\pm$0.015) & 0.697($\pm$0.013)\\
LSTM &  0.759($\pm$0.019) & 0.737($\pm$0.011) & 0.748($\pm$0.020) & 0.722($\pm$0.010) & 0.710($\pm$0.020) & 0.693($\pm$0.014)\\
S2V & 0.789($\pm$0.038) & 0.758($\pm$0.018) & 0.787($\pm$0.040) & 0.753($\pm$0.018) & 0.746($\pm$0.038) & 0.712($\pm$0.020)\\
LSTM$^{m1}$ & 0.804($\pm$0.019) & 0.766($\pm$0.009) & 0.791($\pm$0.021) & 0.753($\pm$0.011)& 0.760($\pm$0.019) & 0.701($\pm$0.010)\\ 
SPAMS$^{rnn}$ & 0.779($\pm$0.045) & 0.759($\pm$0.016) & 0.769($\pm$0.046) & 0.758($\pm$0.017) & 0.725($\pm$0.040) & 0.717($\pm$0.016)\\
SPAMS$^{nc}$ & 0.831($\pm$0.025) & 0.786($\pm$0.013) & 0.827($\pm$0.024) & 0.786($\pm$0.014) & 0.789($\pm$0.026) & 0.738($\pm$0.014)\\
SPAMS & 0.887($\pm$0.024) & 0.823($\pm$0.012) & 0.873($\pm$0.020) & 0.818($\pm$0.012) & 0.822($\pm$0.022) & 0.757($\pm$0.014)\\
\hline
\end{tabular}
\label{per_crop}
\end{table*}

\section{Experiments}
\label{sec:exp}
In this section, we present our evaluation of the proposed method on two real-world datasets. We first introduce the involved baseline methods:


\noindent\underline{Artificial Neural Networks (ANN):} 
We concatenate sequential data and apply ANN, which is a static baseline.


\noindent\underline{Random Forest (RF):} As another static baseline, RF is also applied on  data concatenation. RF has been widely utilized for classifying \textit{remote sensing data}.

\noindent\underline{Hidden Markov Support Vector Machine (SVM$^{hmm}$)~\cite{altun2003hidden}:} This baseline combines SVM and Hidden Markov Model (HMM), and is widely used for sequential labeling. 
We utilize the features at each time step as input and 
copy the sequence label to all the time steps. The prediction is based on majority voting from all the time steps.

\noindent\underline{Long-short Term Memory (LSTM):} 
Similar with SVM$^{hmm}$, we take the features at each time step and copy the sequence label to each time step when applying a traditional LSTM. 

\noindent\underline{Sequence to Vector (S2V)~\cite{li2015improving}:} This baseline is originally designed for \textit{EEG recognition}, and is very relevant to SPAMS 
in that a sliding window is utilized to capture several frequent patterns. Then the sequential data are translated into a vectorized representation using word2vec technique~\cite{mikolov2013distributed}. Finally an SVM classifier is applied on the obtained representation.  

\noindent\underline{LSTM many to one (LSTM$^{m1}$):} We utilize the sliding window and the LSTM structure as described in SPAMS. However, instead of using multi-instance learning approach, this baseline directly utilize the many-to-one LSTM output structure, where we set the sequence label to be the label of the last time window. 

\noindent\underline{Recurrent Neural Networks variant (SPAMS$^{rnn}$):} In this baseline, we replace the LSTM structure in SPAMS with traditional Recurrent Neural Networks (RNN).

\noindent\underline{No context variant (SPAMS$^{nc}$):} This baseline is a variant of SPAMS without using context information.

\subsection{Cropland Mapping}
In this experiment, we implement SPAMS to distinguish between corn and soybean in Minnesota, US. To populate the input sequential features, we utilize MODIS multi-spectral data, 
collected by MODIS instruments onboard NASA's satellites. MODIS data are available for every 8 days and have 46 time steps in a year. At each time step, MODIS dataset provides reflectance values on 7 spectral bands (620-2155 nm) for every location. In this test, we 
took 5,000 locations in Minnesota State for each of corn class and soybean class in 2014 and 2016. 
The ground-truth information on these two classes is provided by USDA crop layer product~\cite{usda}. This task is challenging in agricultural research mainly for two reasons: 1) corn and soybean frequently look similar with each other in most dates of a year, 2) each MODIS location is in 500 m spatial resolution and may contain multiple crop patches, likely introducing noisy features, and 3) the remote sensing data are likely to be disturbed by natural variables (cloudes, smoke, etc.) and other noise factors. 

\begin{figure} [!t]
\centering
\subfigure[]{ \label{fig:a}
\includegraphics[width=0.31\columnwidth]{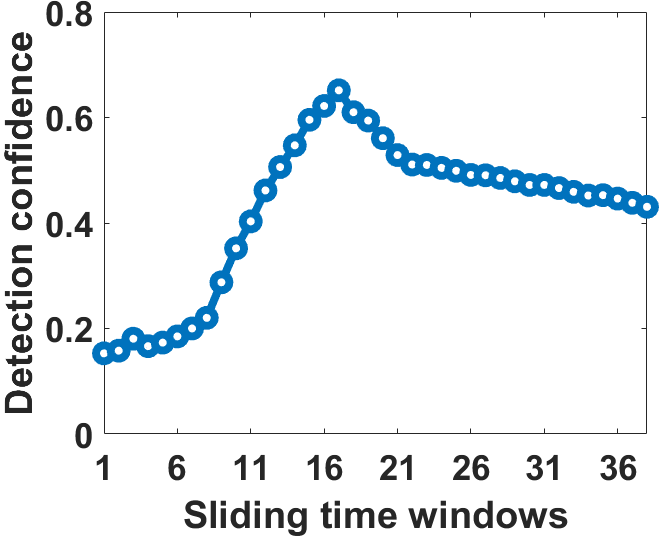}
}\hspace{-.07in}
\subfigure[]{ \label{fig:b}
\includegraphics[width=0.31\columnwidth]{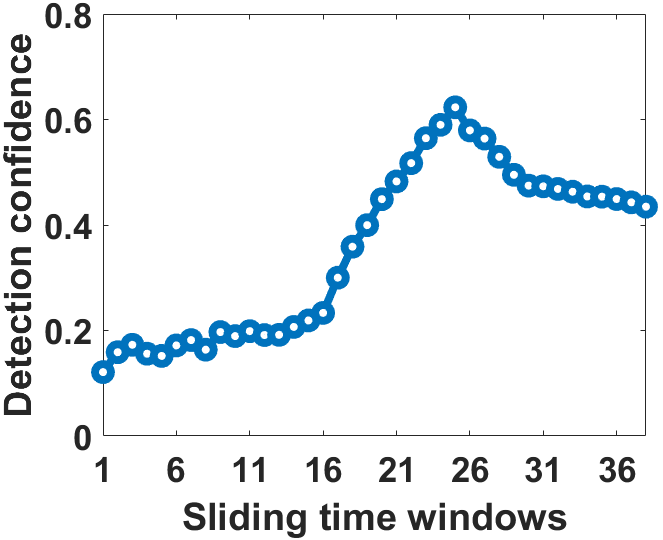}
}\hspace{-.07in}
\subfigure[]{ \label{fig:b}
\includegraphics[width=0.31\columnwidth]{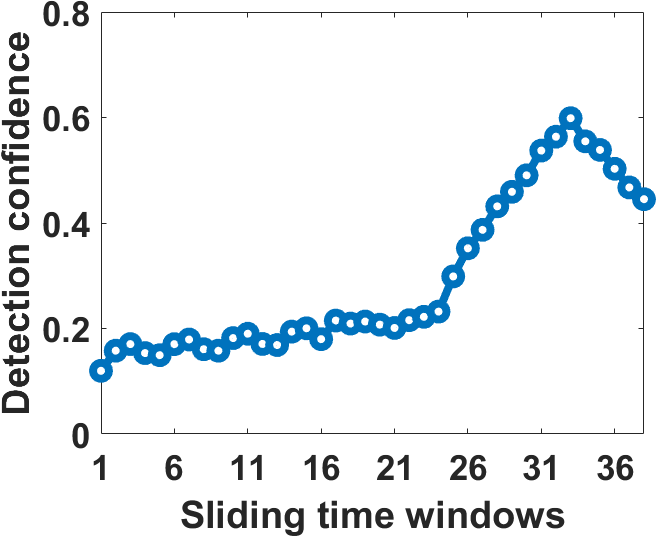}
}\vspace{-.15in}
\caption{The detection confidence on corn samples with delay of (a) 0, (b) 8, and (c) 16 time steps.}
\label{shifting}
\end{figure}

We randomly select 40\% locations from east Redwood, Minnesota and utilize their sequential features in 2016 as training data, and take another 10\% as validation set. Then we conduct three groups of tests: 1) We test on the subset of remaining locations in 2016 which are in the same region with training data (\textit{R1}). 2) We test on the subset of remaining samples in 2016 that are located in different regions with training data (\textit{R2}). 3) We conduct a cross-year test on the data acquired from 2014 using the learned models from 2016. It is noteworthy that the planting time differs between these two years because of the weather conditions in Minnesota. In this test, we set $w=5$, which is sufficiently long to cover an informative period in crop growing process. The selected $l_k$ values are 4 and 5 for corn and soybean class. We present sensitivity test on some hyper-parameters in appendix. The context information is provided based on geo-spatial separation. 

\begin{figure} [!t]
\centering
\subfigure[]{ \label{fig:a}
\includegraphics[width=0.35\columnwidth]{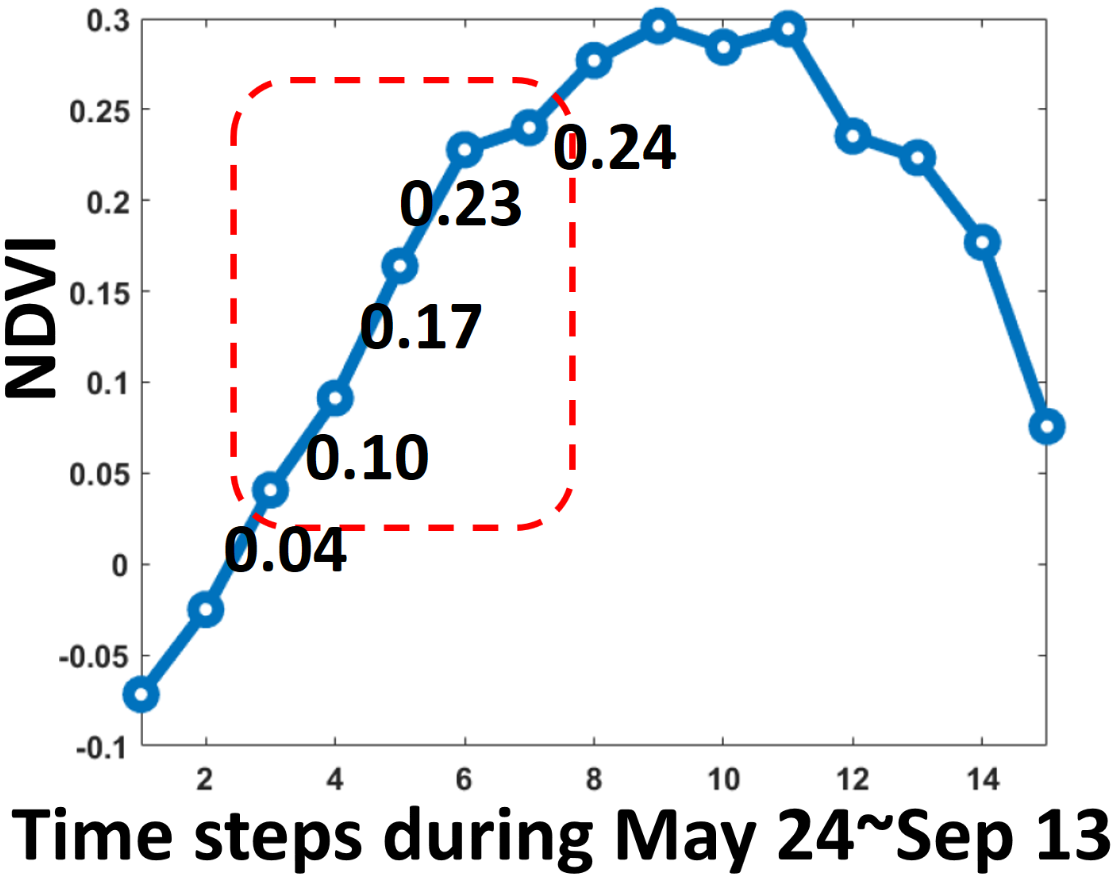}
}
\subfigure[]{ \label{fig:b}
\includegraphics[width=0.35\columnwidth]{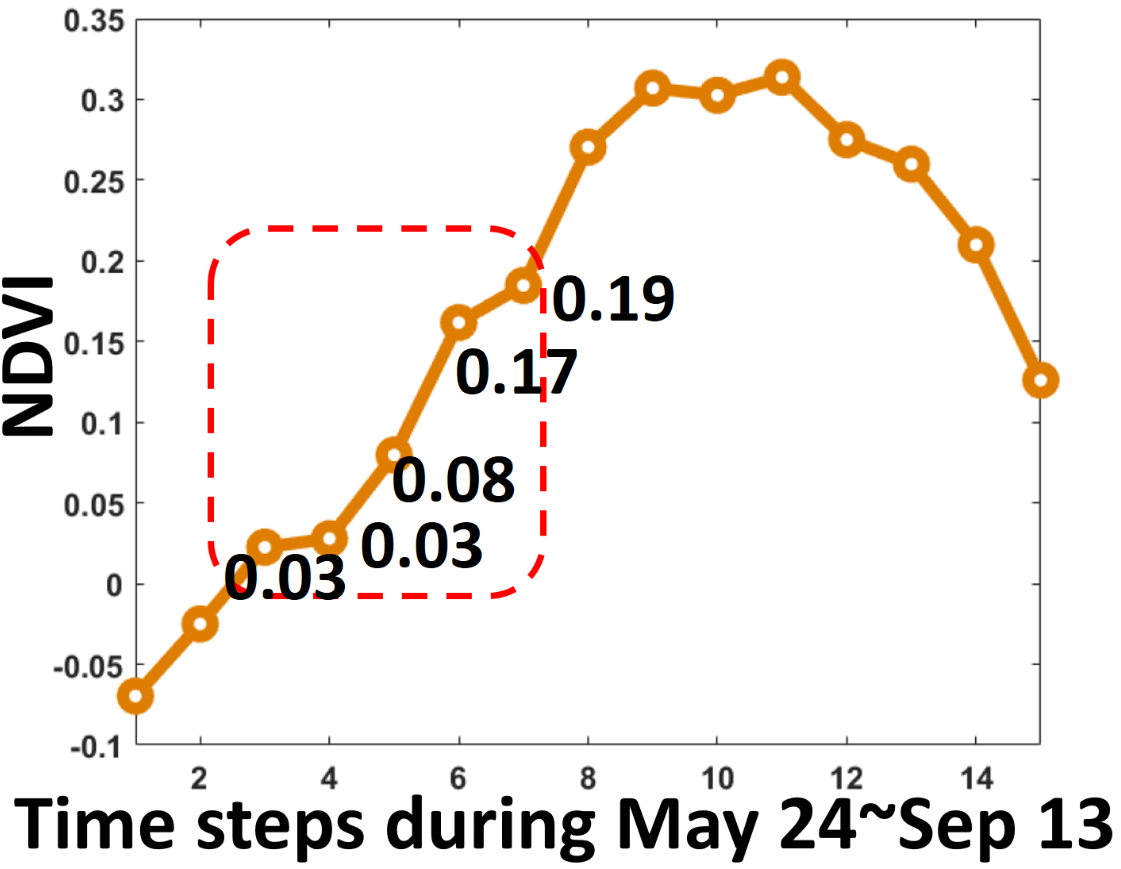}
}\vspace{-.15in}
\caption{The average NDVI (greenness level) series for (a) corn and (b) soybean during May 24$\sim$Sep 13. The blocked part indicates the period Jun 9$\sim$Jul 11.}
\label{dis_pattern}
\end{figure}

\begin{figure*} [!h]
\centering
\subfigure[]{ \label{fig:a}
\includegraphics[width=0.25\columnwidth]{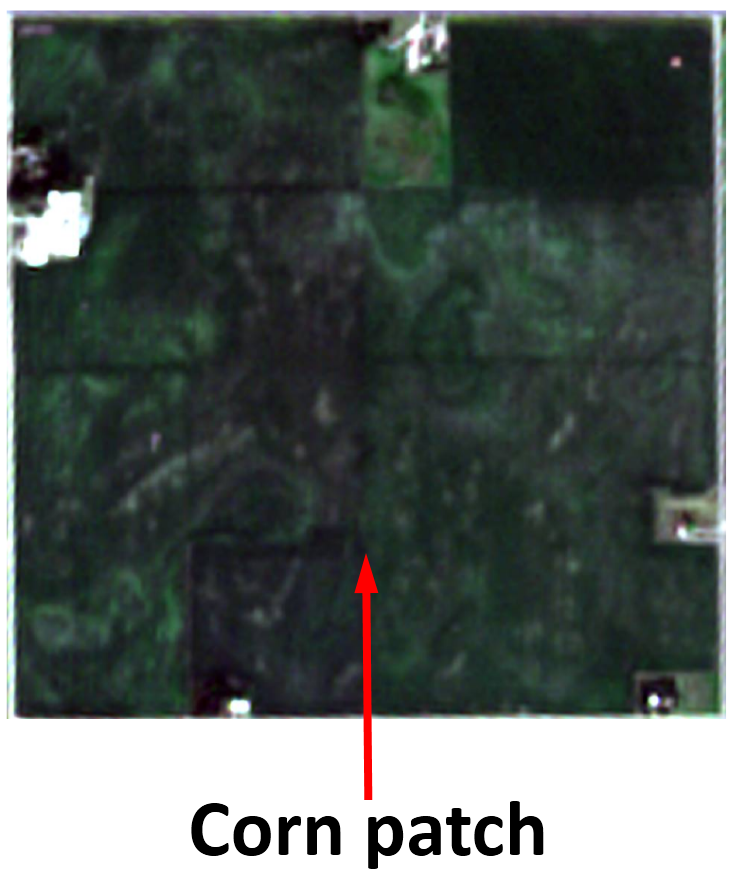}
}
\subfigure[]{ \label{fig:b}
\includegraphics[width=0.245\columnwidth]{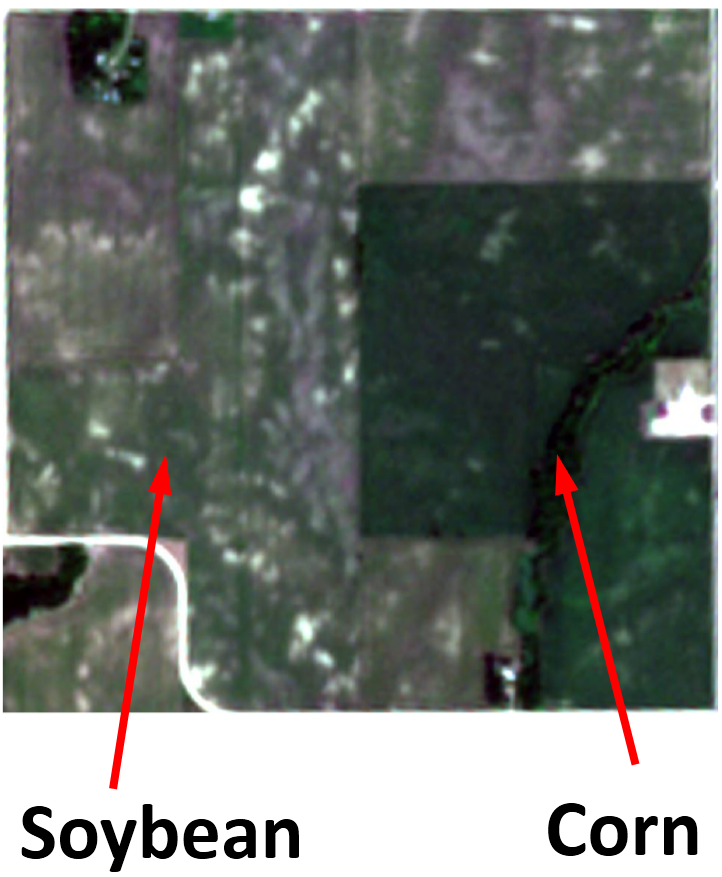}
}
\subfigure[]{ \label{fig:a}
\includegraphics[width=0.27\columnwidth]{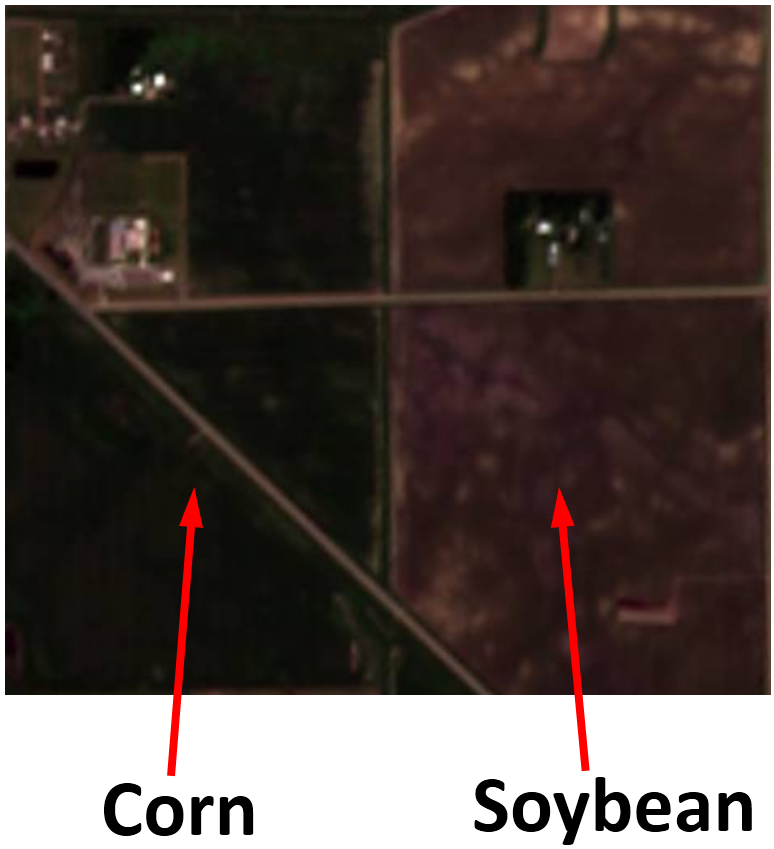}
}
\subfigure[]{ \label{fig:b}
\includegraphics[width=0.226\columnwidth]{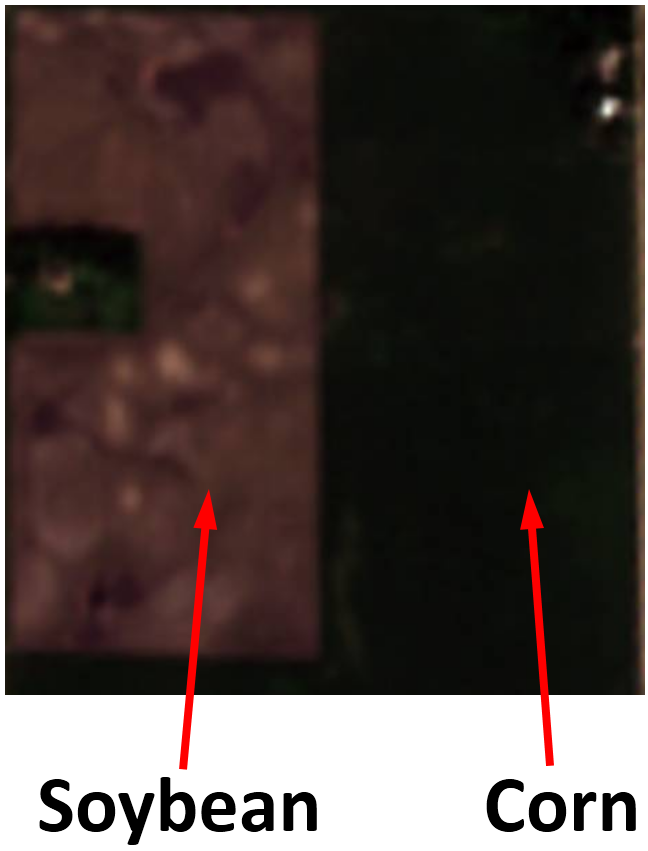}
}
\subfigure[]{ \label{fig:b}
\includegraphics[width=0.476\columnwidth]{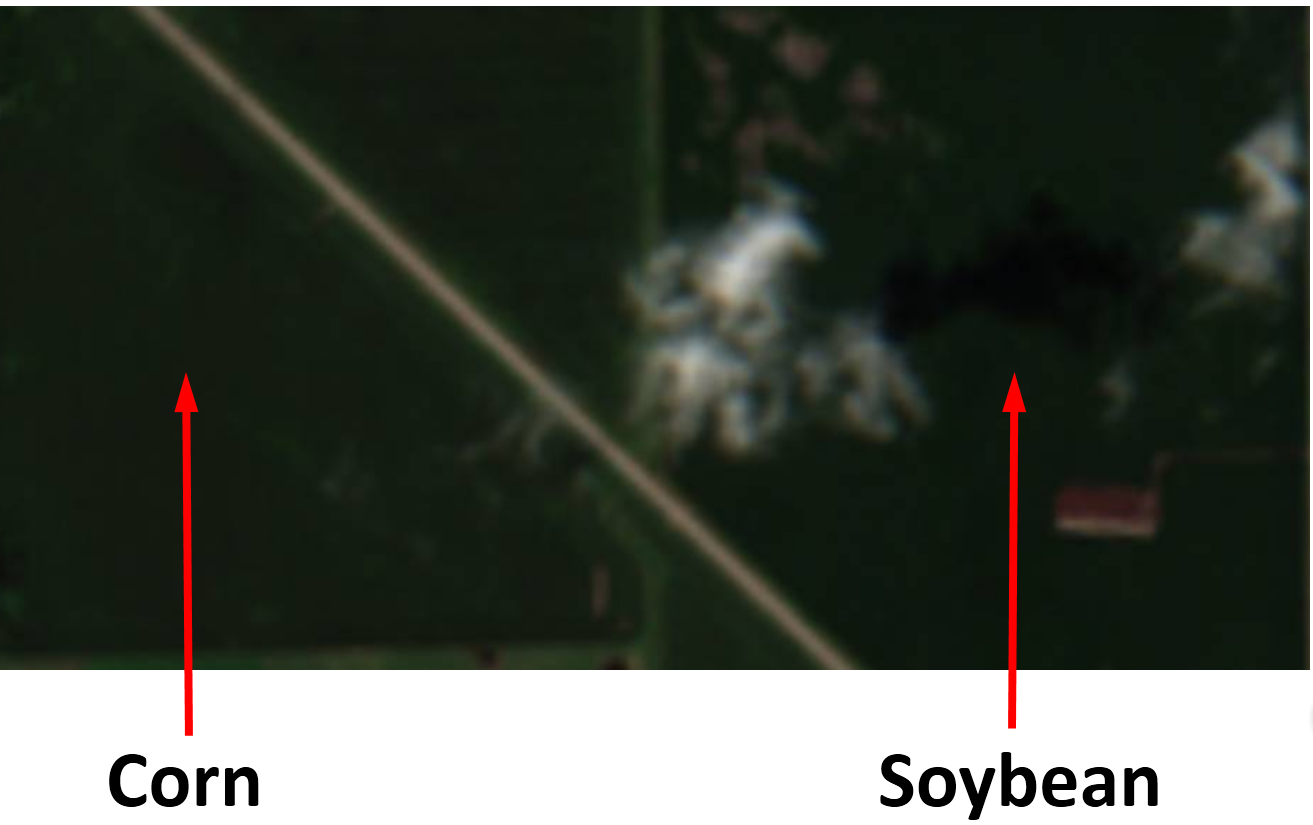}
}\vspace{-.15in}
\caption{Sentinel-2 satellite images. (a)-(d) Cropland patches with corn and soybean on Jun 23, 2016. (e) Another cropland patch taken on Aug 06, 2016.}
\label{crop_ex}
\end{figure*}

We repeat the experiment with random initialization and random selection of training set (from the same region). Then we report the average performance of each test 
in Table~\ref{per_crop}. It can be observed that both static methods (ANN and RF) and sequential methods on time step level (SVM$^{hmm}$ and LSTM) give unsatisfactory performance. By comparing SPAMS$^{rnn}$ and SPAMS, we conclude that long-term dependencies are important for extracting patterns from yearly multi-spectral sequence.  
S2V does not perform as well as SPAMS since many frequent patterns are noisy fluctuation or common patterns for both classes, thus do not contain enough discriminative power. The improvement from LSTM$^{m1}$ and SPAMS$^{nc}$ to SPAMS shows that both multi-instance learning and context information are helpful for the classification.

Table~\ref{per_crop} also shows a decrease of AUC and F1-score for the cross-year performance compared to the performance in 2016. 
This is mainly due to two reasons. First, the planting time of 2014 is in ahead of 2016, and thus a successful classification requires the method to automatically detect such shifting patterns. Second, the collected multi-spectral features vary across years due to environmental variables, such as precipitation, sunlight, etc. Nevertheless, it can be seen that SPAMS still produces a reasonable cross-year detection, which stems from its capacity in capturing shifting patterns. Since domain experts are extremely interested in annual cropland mapping using limited samples, we present more cross-year testing results in the appendix. 

To explicitly show this capacity, we test SPAMS on a synthetic sequence set with shifting patterns. Specifically, we manipulate corn samples by delaying the growing season by 0, 8, and 16 time steps, respectively. In this way, we create three groups of corn samples for testing. We still use the learned SPAMS model from the original training set. In Fig.~\ref{shifting}, we show the detection confidence (i.e. $p^t$ values) on these three groups of test data. It can be clearly seen that SPAMS is capable of locating the most discriminative periods for each group.

To verify that SPAMS indeed detects the discriminative information in multi-variate sequence, we check the obtained top-2 most informative periods in 2016 (results on more detected periods are provided in appendix). The first detected informative period is from Jun 9 to July 11, displayed as the blocked part in Fig.~\ref{dis_pattern}. 
We can find that corn shows higher greenness level in this detected period. Fig.~\ref{crop_ex} (a)-(d) show some corn and soybean patches in four example regions using high-resolution Sentinel-2 images on Jun 23, which confirm that corn patches show a higher greenness level than nearby soybean patches. When applying the learned model to 2014, we can detect the period May 25 to Jun 26 as an informative period, which conforms to the fact that the planting in 2014 starts earlier than 2016.


The second detected informative period in 2016 is from Jul 19 to Aug 20. During this period, both corn and soybean samples show very high greenness level and therefore it is difficult to distinguish between them from either Normalized Difference Vegetation Index (NDVI) series or high-resolution RGB images (e.g. the Aug 06 Sentinel-2 image shown in Fig.~\ref{crop_ex} (e)). Here to verify that this period is indeed an informative period using multi-spectral features, we only use the multi-spectral features from Jul 19 to Aug 20 to train and test a simple ANN model, which produces AUC and F1-score of 0.829 and 0.778, respectively. It is noteworthy that the ANN baseline using the entire sequence only leads to a performance of 0.717 (AUC) and 0.711(F-1 score) according to Table~\ref{per_crop}. This improvement demonstrates that SPAMS has potential to detect the informative or discriminative period of full multi-spectrum, which can not be easily observed by human experts.


\begin{figure} [!t]
\centering
\subfigure[]{ \label{fig:b}{}
\includegraphics[width=0.433\columnwidth]{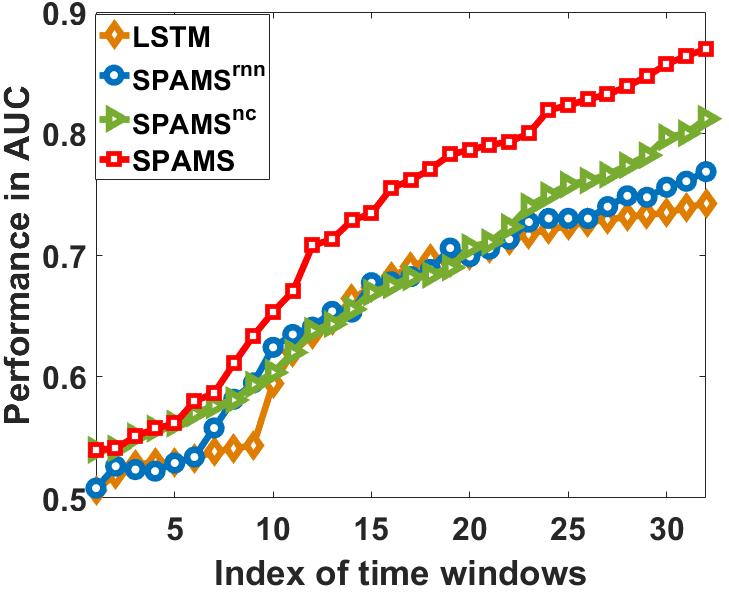}
}
\subfigure[]{ \label{fig:b}{}
\includegraphics[width=0.435\columnwidth]{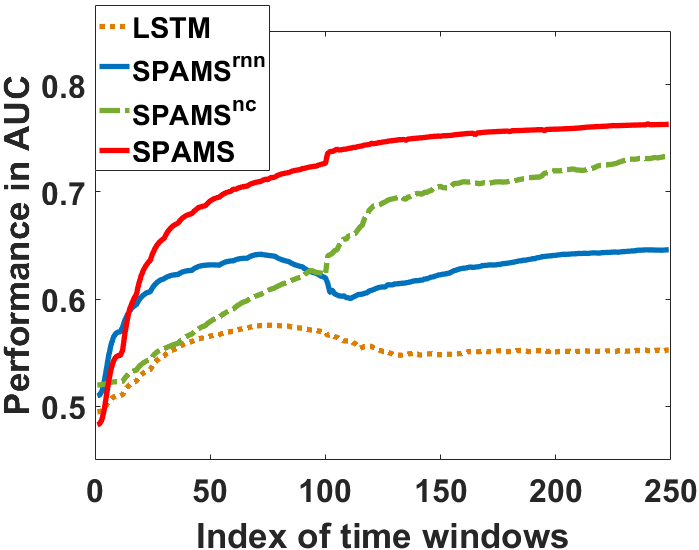}
}\vspace{-.15in}
\caption{The early-stage prediction performance in (a) crop mapping and (b) affective state recognition using sequential data by the end of sliding time window.}
\label{crop_early}
\end{figure}
Finally, we evaluate the performance of SPAMS in early stage prediction. In Fig.~\ref{crop_early} (a), we show the performance of LSTM, SPAMS$^{rnn}$, SPAMS$^{nc}$ and SPAMS given only the sequential data by the end of sliding time windows. We can observe that SPAMS quickly attains higher AUC than other baseline methods shortly after the start of growing season.  




\subsection{Affective state recognition}     
Besides the cropland classification, we validate the proposed method in affective state recognition using collected electroencephalogram (EEG) data. Specifically, we conduct experiment on DEAP EEG dataset~\cite{koelstra2012deap}, where 40-channel EEG data are recorded for 32 participants. Every participant is asked to watch 40 online one-minute videos during EEG collection. We aim to utilize the EEG recordings to classify whether a participant likes a video or not. Before we test each method, we first preprocess the EEG data by downsampling and standardization.

The main challenge in this task is the lack of large volume of training samples. Many existing methods can easily lead to overfitting in this task. According to previous study~\cite{xun2016detecting}, 
we set the length of sliding window to be 1s to cover neural activities, and set the step size of sliding window as 0.25s. The selected $l_k$ values are 8 and 12 for positive and negative class. We present sensitivity test on some hyper-parameters in appendix. 
In our implementation, we utilize the provided experiment settings information as context information. If a set of EEG sequences are recorded when the involved participants are watching the same video, these sequences should belong to the same context.  

We partition the data using 40\% as training data, 10\% as validation set, and 50\% as test set. In Table~\ref{per_eeg}, we show the performance of each method. We can observe that static methods perform poorly because of overfitting. Moreover, compared to SVM$^{hmm}$ and LSTM, the methods that utilize time windows achieve a better performance. This stems from the fact that affective states are reflected by temporal patterns rather than by any single time step. Furthermore, the comparison between SPAMS$^{nc}$ and SPAMS confirms that context information is helpful in locating discriminative time windows and improving classification performance. In general, the prediction accuracy is not as high as cropland mapping task, which is mainly because the participants behave quite differently with each other and we are not provided with sufficient training data.


\begin{table}
\footnotesize
\centering
\caption{Performance($\pm$standard deviation) of each method in affective state recognition. }
\begin{tabular}{l|cc}
\hline
\textbf{Method} & \textbf{AUC$\pm$std} & \textbf{F1$\pm$std} \\ \hline 
ANN & 0.555($\pm$0.021) & 0.624($\pm$0.013)\\  
RF & 0.509($\pm$0.017) & 0.620($\pm$0.014)\\
SVM$^{hmm}$ & 0.569($\pm$0.010) & 0.643($\pm$0.007)\\
LSTM &  0.560($\pm$0.020) & 0.654($\pm$0.012)\\
S2V & 0.705($\pm$0.036) & 0.714($\pm$0.017)\\ 
LSTM$^{m1}$ & 0.709($\pm$0.018) & 0.711($\pm$0.012)\\ 
SPAMS$^{rnn}$ & 0.646($\pm$0.046) & 0.712($\pm$0.019)\\
SPAMS$^{nc}$ & 0.733($\pm$0.021) & 0.722($\pm$0.016)\\
SPAMS & 0.759($\pm$0.020) & 0.746($\pm$0.015)\\
\hline
\end{tabular}
\label{per_eeg}
\end{table}

In Fig.~\ref{crop_early} (b), we show the prediction performance using EEG data by the end of sliding time window. We can observe that SPAMS quickly approaches AUC 0.7 at around 60$^{th}$ time window ($\sim$15s in one-minute video).


Then we apply the learned model to EEG recordings and validate the detected most discriminative time periods for each recording. 
For this validation, we utilize the provided frontal facial videos for the first 22 participants provided by DEAP dataset. For example, if the detected most discriminative time period for ``like" class is around $t$ s, and we observe a smile at the same time. Then we confirm that this detection is correct. 
\begin{table}[!h]
\footnotesize
\newcommand{\tabincell}[2]{\begin{tabular}{@{}#1@{}}#2\end{tabular}}
\centering
\caption{The validation of informative periods (for the first 5 test videos) using synchronized facial videos. We show the informative period (1s) and the fraction of participants showing a relevant expression (rel-expr) to all participants with facial videos (partic).} 
\begin{tabular}{c|l|c|c}
\hline
\textbf{Video} & \textbf{label} & \textbf{time (s)} & \textbf{rel-expr/partic} \\ \hline 
\multirow{2}{*}{1} & like & 19 & 7/11\\  
 & not like & 44 & 5/11\\ \hline
\multirow{2}{*}{2} &  like & 39 & 6/10 \\
 & not like & 14 & 6/12 \\ \hline
\multirow{2}{*}{3} &  like & 11 & 5/9 \\
 & not like & 41 & 6/12 \\ \hline
\multirow{2}{*}{4} &  like & 23 & 6/13 \\
 & not like & 6 & 5/9 \\ \hline
\multirow{2}{*}{5} &  like & 2 & 5/9\\
 & not like & 56 & 6/12 \\ \hline
\end{tabular}
\label{int_eeg}
\end{table}

Since it is time-consuming to check each individual trial, in this test we compute the discriminative time period for each video using the average temporal profile. Then for each video, we manually check how many participants out of all participants have a relevant facial expression around the detected discriminative time period (allowing a delay of 2s). The relevant facial expressions include moves of lips, eyebrow, nose, eyeballs, etc. In Table~\ref{int_eeg}, we report only first 5 videos due to space limit. According to our study on all 40 videos, the average fraction values for ``like" and ``unlike" are 0.569 and 0.495. Since most participants show very few facial expressions during entire process of EEG recording, these fraction values can clearly confirm that SPAMS detects the discriminative time steps that are relevant to the classification.





\section{Related Work}
\label{sec:rw}
Sequential data have been widely studied in web browsing analysis, bioinformatics, remote sensing~\cite{adamo2012data,jianya2008review}. 
Over the past decade, a predominant portion of sequential data modeling/prediction has been proposed based on HMM and its variants~\cite{rabiner1989tutorial,altun2003hidden,dietterich2002machine}. 
Yet, despite its success, HMM has several limitations. First, the conditional independence assumptions are often restrictive. Second, it lacks the power of capturing non-linear transitions. 

With the development of deep learning, RNN model~\cite{graves2013speech} has been ubiquitously implemented.  
It is however a challenge 
to train RNN efficiently for vanishing gradient problem~\cite{bengio1994learning}. 
The most recent RNN variant towards modeling a complicated transitional distribution is the Recurrent Temporal Restricted Boltzmann Machine (RTRBM)~\cite{sutskever2009recurrent}, which is a straightforward combination of RNN and RBM. 
Unfortunately, RTRBM still suffers from the drawbacks of RNN~\cite{sutskever2009recurrent}. 
To solve the vanishing gradient problem, researchers have proposed extensions of RNN, including LSTM and Gated Recurrent Unit (GRU)~\cite{chung2014empirical}, which has proven to be effective in memorizing long-term dependencies.

These sequence classification methods are frequently combined with specific data 
properties 
to improve the performance~\cite{liu2016unified,zhou2016pattern}. 
In particular, the sequential patterns have drawn great attention
~\cite{yuan2011discriminative,xun2016detecting,batal2012mining,lee2011mining}. For instance,~\cite{xun2016detecting} detects frequent patterns from EEG sequences and then transforms these patterns as input to a classification model. However, this method may capture irrelevant patterns or noise factors with limited discriminative information. 
Another work~\cite{yuan2011discriminative} 
focuses on handling dynamic video backgrounds in searching reoccurences of predefined discriminative pattern, but cannot be applied to extract discriminative patterns.


In this work, we combine LSTM with MIL in sequence classification. 
There exist only a few works that utilize MIL in sequential data~\cite{xu2011ensemble,hu2009action} and they are not relevant to discriminative pattern detection. In contrast, we utilize LSTM-based structure to model the sequential structure within each bag and then utilize an MIL approach to detect discriminative patterns.


\section{Conclusions}
In this paper, we propose a novel method for sequence classification by discovering the discriminative patterns within the sequence. We utilize a sliding window to capture the discriminative pattern and combine it with LSTM to incorporate temporal dependencies. Besides, an MIL structure is introduced to detect the discriminative period, which also provides real-world interpretation to prediction results in applications. According to experimental results, SPAMS outperforms multiple baselines in sequence classification. Besides, 
the successful detection of discriminative periods is extremely valuable for scientific domain research, which used to heavily rely on simple classification methods with  hand-crafted temporal features 
from the entire sequence. 
In this way, SPAMS has potential to contribute to a large class of inter-disciplinary works between machine learning and scientific domain research.
\bibliographystyle{abbrv}

\bibliography{IEEEabrv}

\end{document}